\documentclass[11pt]{article}

\usepackage[utf8]{inputenc}
\usepackage[T1]{fontenc}
\usepackage{lmodern}
\usepackage{amsmath,amssymb}
\usepackage{graphicx}
\usepackage{booktabs}
\usepackage{enumitem}
\usepackage{tikz}
\usepackage{hyperref}
\usepackage{geometry}
\usepackage{color}

\title{Predicting Talent Breakout Rate using Twitter and TV data\thanks{%
Originally presented at the 34th Annual Conference of the Japanese Society for Artificial Intelligence (JSAI 2020), paper ID 1K3-ES-2-02.\\
Corresponding authors: \texttt{hiroyuki.fukuda@dentsu.co.jp}, \texttt{batsaikhan.bilguun@dentsu.co.jp}.%
}}

\author{Batsaikhan Bilguun$^{1,2}$ \and Hiroyuki Fukuda$^{1,2}$\\[0.5em]
$^1$Dentsu Inc.\\
$^2$Data Artist Inc.}

\date{} 

\def\BibTeX{{\rm B\kern-.05em{\sc i\kern-.025em b}\kern-.08em%
 T\kern-.1667em\lower.7ex\hbox{E}\kern-.125emX}}
\def\JBibTeX{\leavevmode\lower .6ex\hbox{J}\kern-0.15em\BibTeX}
\def\LaTeXe{\LaTeX\kern.15em2$_{\textstyle\varepsilon}$}

\begin{document}
\maketitle

\begin{abstract}
Early detection of rising talents is of paramount importance in the field of advertising. In this paper, we define a concept of talent breakout and propose a method to detect Japanese talents before their rise to stardom. The main focus of the study is to determine the effectiveness of combining Twitter and TV data on predicting time-dependent changes in social data. Although traditional time-series models are known to be robust in many applications, the success of neural network models in various fields (e.g.\ Natural Language Processing, Computer Vision, Reinforcement Learning) continues to spark an interest in the time-series community to apply new techniques in practice. Therefore, in order to find the best modeling approach, we have experimented with traditional, neural network and ensemble learning methods. We observe that ensemble learning methods outperform traditional and neural network models based on standard regression metrics. However, by utilizing the concept of talent breakout, we are able to assess the true forecasting ability of the models, where neural networks outperform traditional and ensemble learning methods in terms of precision and recall.
\end{abstract}

\section{Introduction}

Twitter is one of the biggest social networking platforms in Japan. Millions of tweets are produced on a daily basis. As a result, to utilize the rich information base of the platform, researchers have conducted many compelling experiments. One of the research topics is trend prediction. Although the definition of trend varies based on context, the most general interpretation is predicting future behavior of a time dependent system given historical data. In this study, we build upon the fundamental properties of trend prediction to find Japanese entertainment industry talents on their path to prominence. With automatic detection of rising talents, companies get the opportunity to utilize various advertising opportunities such as first-mover advantage to promote and collaborate with the rising talents.

The main contribution of our paper lies in determining the combined effect of Twitter and TV data in predicting trends. To deal with the highly volatile and context-specific trend patterns of the Twitter platform, we incorporate mass media signals (TV) to aid the low signal-to-noise ratio of the Twitter data set.

\section{Related Work}

One of the interesting approaches in detecting trends on Twitter is the latent source classification model, proposed in \cite{George2013}. The method evaluates changes over time in the number of tweets about each new topic to the changes over time of every sample in the training set. Specifically, samples whose statistics resemble those of the new topic are given more weight in predicting whether the new topic will trend or not. In effect, each sample ``votes'' on whether the new topic will trend, but some samples' votes count more than others'. The weighted votes are then combined, giving a probabilistic estimate of the likelihood that the new topic will trend.

In \cite{Manish2012}, researchers have tried comparing traditional statistical methods with ML methods based on Twitter data. One of the interesting aspects of the study was the transformation of regression models' predictions into classification labels. With this they were able to compare multiple regression and classification techniques to find the best modeling approach for their data set.

In \cite{Zhang2013}, by creating content and context (user-specific) features from tweets, the researchers investigated important factors and effective models for learning trends on Twitter. They have found out that non-linear models are significantly better than their linear peers, which is due to the complex information diffusion processes in large social networks.

Prior works have explored a variety of modeling techniques with different objectives, granularity of time units, and number of engineered features. The main contribution of our work lies in establishing a baseline for predicting rising talents with the combination of Twitter and TV data by comparing traditional forecasting methods with ML methods. We introduce a simple framework to find the best modeling approach for the social network data and evaluate our results on multiple metrics.

\section{Data}

A list of 5000 Japanese talents was used to create the training data set. The list contains both prominent and rising talents from various entertainment fields. The statistics of the data set are shown in Table~\ref{tab:stats}.

\textbf{Twitter.} For research purposes, we have used Twitter's ``Sample real time Tweets'' API and recorded tweets on a daily basis beginning from March 2019 till January 2020. Approximately 700{,}000 tweets are queried each day.

A simple but powerful feature is the number of mentions. Counting how many times a talent was mentioned in a day can provide substantial information. Although we can create a variety of context and user dependent features, to quickly iterate through models and understand the combined predictive power of Twitter and TV data, we have concentrated on number of mentions. Considering our objective, we have removed outlier talents with too low (total mentions $\leq 10$) and too high (total mentions $\geq 5000$) mentions.

Finding the right granularity for time series data is crucial. Large time periods might hinder our capabilities of learning interesting signals in the data, while too short time periods might result in large and noisy data sets \cite{Gonzales2018}. Since we are predicting three months in advance and the time span of the data set is 10 months, we have transformed our data from daily to weekly counts. Although we have chosen weekly granularity, it is possible for data sets with finer granularity (daily, hourly) to perform better.

\textbf{TV.} In terms of TV, we have used our in-house data set. The data is produced by recording each talent's occurrence on Japanese TV programs. Out of 5000 talents in our list, approximately half of the talents had a history of being mentioned on TV. Since traditional forecasting methods (vector autoregressions) are trained for each talent (which means there is no sense training models on talents without TV data), to conduct proper comparison of the models using same test data, we have performed our studies on talents with TV data.

Compared to TV industries in other parts of the world, the Japanese TV industry is still effective for reaching the masses. Today, virtually all Japanese households have one or more TVs, and TV broadcasting is extremely important as a vehicle for advertisement, as a source for breaking news, and as a means to access major sports and national events \cite{Oku2019}. For these reasons, we have decided to incorporate TV signals into the Twitter dataset.

\begin{table}[t]
\centering
\caption{Twitter and TV statistics.}
\label{tab:stats}
\begin{tabular}{lr}
\toprule
\textbf{Talents with Twitter data} & 4106\\
\textbf{Talents with TV data} & 2492\\
\textbf{Total time span} & 45 weeks\\
\textbf{Average Twitter mentions per talent} & 662\\
\textbf{Average TV mentions per talent} & 52\\
\bottomrule
\end{tabular}
\end{table}

\section{Framework}

\subsection{Objective}

We define our task as a multi-feature time series regression. As the modeling objective we are predicting talents' future 3rd month's average Twitter count. Sample input and output training data is illustrated below, where $m$ specifies months:

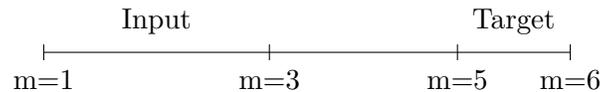
\begin{center}
\begin{tikzpicture}
\draw (0,0) -- (7,0);
\foreach \x in {0,3,5.5,7}
\draw (\x cm,3pt) -- (\x cm,-3pt);
\draw (0,0) node[below=3pt] {m=1} node[above=3pt] {};
\draw (3,0) node[below=3pt] {m=3} node[above=3pt] {};
\draw (1.5,0) node[above=3pt] {Input};
\draw (5.5,0) node[below=3pt] {m=5} node[above=3pt] {};
\draw (7,0) node[below=3pt] {m=6} node[above=3pt] {};
\draw (6.25,0) node[above=3pt] {Target};
\end{tikzpicture}
\end{center}

Depending on the size of the training data set and the objective, we recommend experimenting with different training and target time spans. According to our results, using 3 months for training and predicting future 3rd month's average counts produced stable results.

\subsection{Autoregressive models}

The most common method in traditional time series forecasting is the Autoregressive Integrated Moving Average model (ARIMA). Its multivariate versions are Vector Autoregressions (VAR) and Vector Autoregressions with Moving Average (VARMA) models \cite{George2008}. For this study we have used the VAR and VARMA models with two endogenous variables (Twitter and TV), no exogenous variable and no constant term.

There are two distinct parameters we have to specify for VAR and VARMA methods. First is AR --- the autoregressive part of the model. It determines how many past periods we consider when making a decision. Second is MA --- the moving average part of the model. It allows for stable measurement of errors. One of the essential parts of using VAR and VARMA models is the necessity of carefully selecting the AR and MA parameters. The large number of series makes it difficult to hand-pick the best parameters for each series. However, one solution is to select the AR and MA parameters of a model based on its validation set performance as shown in the figure below, where $w$ specifies weeks.

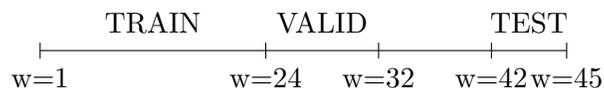
\begin{center}
\begin{tikzpicture}
\draw (0,0) -- (7,0);
\foreach \x in {0,3,4.5,6,7}
\draw (\x cm,3pt) -- (\x cm,-3pt);
\draw (0,0) node[below=3pt] {w=1} node[above=3pt] {};
\draw (3,0) node[below=3pt] {w=24} node[above=3pt] {};
\draw (1.5,0) node[above=3pt] {TRAIN};
\draw (4.5,0) node[below=3pt] {w=32} node[above=3pt] {};
\draw (3.75,0) node[above=3pt] {VALID};
\draw (6,0) node[below=3pt] {w=42} node[above=3pt] {};
\draw (7,0) node[below=3pt] {w=45} node[above=3pt] {};
\draw (6.5,0) node[above=3pt] {TEST};
\end{tikzpicture}
\end{center}

Based on the above figure, we trained the VAR and VARMA models on the first 24 weeks, validated on the next 8 weeks to find the best AR and MA parameters, and calculated the final errors on the last 4 weeks of data.

One of the main requirements of VAR and VARMA models is for the series to be stationary (constant mean, variance and autocorrelation over time). Although it is not perfect, we have used the Augmented Dickey--Fuller unit root test to automatically determine the stationarity of a series \cite{Dickey1979}. To deal with non-stationary series, we have log-differenced the data.

\subsection{Machine learning methods}

\textbf{Ensemble methods.} Ensemble learning methods are considered as the go-to methods in machine learning. Considering no seasonal or long-term trends in the talents' time series data, we can use ensemble learning methods in our application. As representatives of ensemble learning methods, we chose the widely used Random Forests (RF) \cite{Leo2001} and a gradient boosting algorithm, LightGBM (LGBM) \cite{Guolin2017}.

\textbf{Neural network methods.} Despite the many advancements in the field of deep learning, for instance, the introduction of Bidirectional Encoder Representations from Transformers (BERT) in 2018 \cite{Devlin2018}, for the primary purpose of comparing the fundamental learning capabilities of traditional forecasting methods with ML methods, we chose to experiment with Long Short Term Memory networks (although introduced in 1997, one of the most robust neural network models is the LSTM networks) \cite{Sepp1997} and Multi-Layer Neural Networks (MLNN) \cite{David1986}. We plan to experiment with more advanced sequential modeling methods in the near future.

Decision tree algorithms are invariant to monotonic transformations of the features. Therefore, the effect of scaling the input features is considered to be small. On the other hand, neural networks are known to benefit from normalizing the input and output data \cite{Warren1999}. Thus, we have normalized the series by subtracting the mean and dividing by the standard deviation.

Machine learning methods are prominent for their capability to scale. Compared to the training processes of VAR and VARMA models, where we had to train and evaluate one talent at a time, the training of ensemble and neural networks were executed much faster. For hyperparameter optimization of ML methods, we have divided the data into 6 sequential splits, where the first 5 splits were used for training and the last split for evaluation.

\section{Experiments}

In this section, we present the results of 6 different models. The models were trained on Twitter and TV data from 10 March 2019 to 12 October 2019 and evaluated on data from 13 October 2019 to 12 January 2020. We set the time unit as one week. To determine the most suitable modeling algorithm for the data and objective, we evaluated the models on three evaluation metrics:

\begin{enumerate}[itemsep=1pt, topsep=1pt]
  \item Mean Absolute Error (MAE) and Root Mean Squared Error (RMSE);
  \item Precision and recall based on the concept of breakout;
  \item Scalability and interpretability.
\end{enumerate}

\subsection{MAE and RMSE}

\begin{table}[t]
\centering
\caption{Results based on MAE and RMSE metrics. Best methods are highlighted in bold. Ensemble methods' overall performance is higher than traditional and neural network methods. ``-TW'' specifies models trained only on Twitter data to determine the effect of TV signals.}
\label{tab:mae_rmse}
\begin{tabular}{lrr}
\toprule
\textbf{Model} & \textbf{MAE} & \textbf{RMSE}\\
\midrule
VAR         & 17.21 & 78.83\\
VARMA       & 18.35 & 42.67\\
\textbf{RF} & \textbf{12.57} & \textbf{27.76}\\
\textbf{LGBM} & \textbf{11.80} & \textbf{29.17}\\
LGBM-TW     & 11.83 & 29.24\\
LSTM        & 13.11 & 29.58\\
LSTM-TW     & 13.18 & 29.66\\
MLNN        & 15.10 & 32.03\\
\bottomrule
\end{tabular}
\end{table}

Results in Table~\ref{tab:mae_rmse} illustrate the overall accuracy of the models. Ensemble-based models and the LSTM model outperformed the VAR and VARMA models by a considerable amount. Since VAR and VARMA models consider the future values of a series to be a linear function of its past values, the models were prone to making larger errors compared to non-linear models. When it comes to non-linear ML methods, LGBM, due to its superior optimization method of gradient boosting, seems to have achieved the best MAE. On the other hand, with the help of fully grown decision trees that help reduce variance, Random Forests show superior RMSE results. Moreover, we came to the conclusion that the ability to keep or forget information gives an advantage to LSTMs over simple feed-forward NNs. Based on both evaluation metrics, it is evident that ensemble learning methods fit better to the data.

To understand the effect of TV signals, we have trained the top two models without TV data. We can see that there is indeed an improvement in the errors when models are trained with TV data. Considering the relative sparsity of TV signals and the fact that the target variable is calculated using only Twitter counts, we observe that there is indeed a benefit of adding TV signals for our task. However, we estimate the effect is relatively subtle.

\subsection{Precision and recall based on breakout}

We have found that the standard evaluation metrics, such as mean absolute error and root mean squared error, are not capable of capturing the full predictive capability of the models. Therefore, we have decided to define a custom evaluation metric called talent ``breakout'':
\begin{equation}
    \text{Breakout} =
    \begin{cases}
        \text{True}, & \beta / \gamma \geq 1.2, \\
        \text{False}, & \text{otherwise},
    \end{cases}
\end{equation}
where $\beta$ is the future 3rd month's average Twitter count and $\gamma$ is the past 3 months' average Twitter count.

From Equation~(1), we can see that talent breakout is a binary value. It is true for series with the ratio of future 3rd month's Twitter count to past 3 months' Twitter count greater than or equal to 120\%. Once we calculate the predicted and actual breakout rates for each talent, we are able to assess the models' predictions using classification metrics. Particularly, we can use prominent metrics such as precision and recall.

\begin{table}[t]
\centering
\caption{Results based on precision and recall. Best methods are highlighted in bold. Precision was calculated by ordering the results based on the predicted breakout rate and selecting the top 500 talents. Recall was calculated by sorting on the actual breakout rate of the talents.}
\label{tab:breakout}
\begin{tabular}{lrr}
\toprule
\textbf{Model} & \textbf{Precision top 500} & \textbf{Recall top 500}\\
\midrule
VAR        & 46\%   & 35\%\\
VARMA      & 35\%   & 28\%\\
RF         & 39\%   & 52\%\\
LGBM       & 40\%   & 13\%\\
LGBM-TW    & 41\%   & 12.4\%\\
\textbf{LSTM} & \textbf{46\%} & \textbf{44\%}\\
LSTM-TW    & 45.4\% & 45.8\%\\
\textbf{MLNN} & \textbf{44\%} & \textbf{58\%}\\
\bottomrule
\end{tabular}
\end{table}

It was important for us to not only show the top breakout talents on top of the page, but also output as many top breakout talents as possible. Therefore, to rank the models, we have used both precision and recall. According to the breakout results shown in Table~\ref{tab:breakout}, considering the simplicity of the models, VAR and VARMA were stable in terms of precision and recall. When it comes to ensemble learning methods, due to its focus on reducing variance by growing full decision trees, Random Forests have resulted in a considerably high recall rate. On the other hand, compared to the rest of the models where the average number of predicted breakout talents exceeded 500, LightGBM predicted only 116 talents as potential breakouts. We think the low recall rate is caused by LightGBM's tendency to easily overfit when the number of samples is low. Although we have tried several regularization parameters, the results did not change much. Based on all performance metrics, neural network methods outperformed both traditional and ensemble learning methods, and thus were chosen as the most appropriate methods for our data set. When it comes to the effect of TV in terms of breakout rate, there is a slight improvement in the accuracy of the models. Since we can not directly say that TV signals are essential for predicting talent breakouts, we have come to the conclusion that it is necessary to incorporate other types of social media signals (YouTube, Instagram, etc.) in addition to TV.

\subsection{Scalability and interpretability}

\textbf{Scalability.} During our experiments with traditional methods, we trained VAR and VARMA models for each talent series. Although several hours of training in an environment with multiple processing cores is feasible for the production environment, to conduct fast and iterative preliminary experiments, it was necessary to sample the number of talents from the talent list.

\textbf{Interpretability.} In commercial products, having the ability to display the range of possible values creates an incentive to act based on the predictions. Compared to machine learning methods, traditional methods have the inherent capability to produce confidence intervals. This serves as an advantage for algorithms like VAR and VARMA. Furthermore, the deficiency to clearly interpret and explain the predictions of black-box models in ML, especially of neural networks, creates an issue for practitioners to use the methods in production.

\section{Conclusion}

For accurate time series trend prediction, it is necessary to integrate as many data sources as possible. In this study, we have explored the combined effect of Twitter and TV data to predict upcoming talents in Japan. As a result of our experiments, we conclude that, although there is an observed benefit of using TV signals, due to the highly volatile nature of social network data, additional sources of signals are required to improve the accuracy of the models. Moreover, we find that traditional forecasting methods should be given attention before considering ML methods. By tuning the simple hyperparameters of traditional methods (VAR and VARMA), we are able to produce breakout results of same quality compared to those of comprehensively tuned ML methods. However, based on standard regression metrics, due to linear statistical properties, the traditional methods have performed poorly. When it comes to ensemble learning methods, despite their superior MAE and RMSE values, the actual breakout prediction quality of the models was low compared to other methods. On the other hand, based on the overall accuracy of the models, LSTM and Multi-Layer NNs outperformed the traditional and ensemble learning methods. We hope that our results will serve as a contribution to the task of modeling social media trends. In the near future, we are planning to incorporate other social network data sets and explore hybrid modeling methods that are able to output confidence intervals.

\end{document}